\documentclass[lettersize,journal]{IEEEtran}
\usepackage{amsmath,amsfonts}
\usepackage{algorithmic}
\usepackage{algorithm}
\usepackage{array}
\usepackage[caption=false,font=normalsize,labelfont=sf,textfont=sf]{subfig}
\usepackage{textcomp}
\usepackage{stfloats}
\usepackage{url}
\usepackage{verbatim}
\usepackage{graphicx}
\usepackage{cite}
\usepackage{booktabs}
\usepackage{multicol}
\usepackage{multirow}
\usepackage{amsmath}
\usepackage{xspace}
\usepackage[table]{xcolor}
\usepackage{amssymb}
\begin{document}

\title{IEC3D-AD: A 3D Dataset of Industrial Equipment Components for Unsupervised Point Cloud Anomaly Detection}

\author{
	\vskip 1em
	
	Bingyang Guo, \emph{Student Membership}, Hongjie Li, Ruiyun Yu${\dagger}$, \emph{Membership}, Hanzhe Liang, and Jinbao Wang  

	\thanks{
            ${\dagger}$ Corresponding author.
            
		This work was supported in part by the National Key Research and Development Program of China under grant 2024YFB3409202, the Key Research and Development Program of Liaoning Province under grant 2024020969-JH2/1024, and the Fundamental Research Funds for the Central Universities under Grant N2317001.
		
		Bingyang Guo, Hongjie Li, and Ruiyun Yu are with the Software College, Northeastern University, Shenyang, China (e-mail: 2110500@stu.neu.edu.cn, 2271444@stu.neu.edu.cn, yury@mail.neu.edu.cn). 
        
        Hanzhe Liang is with the Shenzhen Audencia Financial Technology Institute, Shenzhen University, Shenzhen, China (e-mail: 2023362051@email.szu.edu.cn). 
        
        Jinbao Wang is with the National Engineering Laboratory for Big Data System Computing Technology, Shenzhen University, Shenzhen, China (e-mail: wangjb@szu.edu.cn).  
	}
}

\markboth{IEEE Transactions on Circuits and Systems for Video Technology}%
{Shell \MakeLowercase{\textit{et al.}}: A Sample Article Using IEEEtran.cls for IEEE Journals}


\maketitle

\begin{abstract}
3D anomaly detection (3D-AD) plays a critical role in industrial manufacturing, particularly in ensuring the reliability and safety of core equipment components. Although existing 3D datasets like Real3D-AD and MVTec 3D-AD offer broad application support, they fall short in capturing the complexities and subtle defects found in real industrial environments. This limitation hampers precise anomaly detection research, especially for industrial equipment components (IEC) such as bearings, rings, and bolts. To address this challenge, we have developed a point cloud anomaly detection dataset (IEC3D-AD) specific to real industrial scenarios. This dataset is directly collected from actual production lines, ensuring high fidelity and relevance. Compared to existing datasets, IEC3D-AD features significantly improved point cloud resolution and defect annotation granularity, facilitating more demanding anomaly detection tasks. Furthermore, inspired by generative 2D-AD methods, we introduce a novel 3D-AD paradigm (GMANet) on IEC3D-AD. This paradigm generates synthetic point cloud samples based on geometric morphological analysis, then reduces the margin and increases the overlap between normal and abnormal point-level features through spatial discrepancy optimization. Extensive experiments demonstrate the effectiveness of our method on both IEC3D-AD and other datasets.
\end{abstract}


\section{Introduction}

\IEEEPARstart{B}{\textbf{ackground}} \textbf{of the IEC3D-AD proposal.} Industrial equipment components (IEC) serve as key connectors to advance industrial foundations and modernize the industrial chain \cite{10969994}. They are essential units for the high-quality growth of the equipment manufacturing industry. Traditional 2D visual acquisition equipments struggle to provide depth information as they cannot accurately depict changes in the shape from various spatial angles \cite{DBLP:journals/ijautcomp/LiuXWLWZJ24,10251020,10542974,1182038}. The development of 3D acquisition technologies, including dual CCD structured light and laser scanners, has substantially mitigated this issue by allowing the extraction of complex 3D data. While numerous studies \cite{DBLP:conf/mm/ChenXLWLWZ23,DBLP:conf/wacv/BergmannS23,DBLP:journals/corr/abs-2311-14897,DBLP:journals/corr/abs-2312-04521,9583886,10360133,10744600,10858074} have investigated the use of 3D data for industrial anomaly detection, the majority of data samples do not originate from actual industrial products, and cannot fully capture the complex situations and subtle defects unique to real industrial environments. Therefore, there is an immediate need to develop a high-precision point cloud dataset that can meet industrial anomaly detection requirements for IEC. What's more, this can also further promote the development of 3D anomaly detection tasks under the challenges of complex shapes and minor defects. To address this need, we propose a dataset named IEC3D-AD, explicitly designed to support the study and enhancement of 3D anomaly detection techniques. 

\begin{figure}
    \centering
    \includegraphics[width=0.9\linewidth]{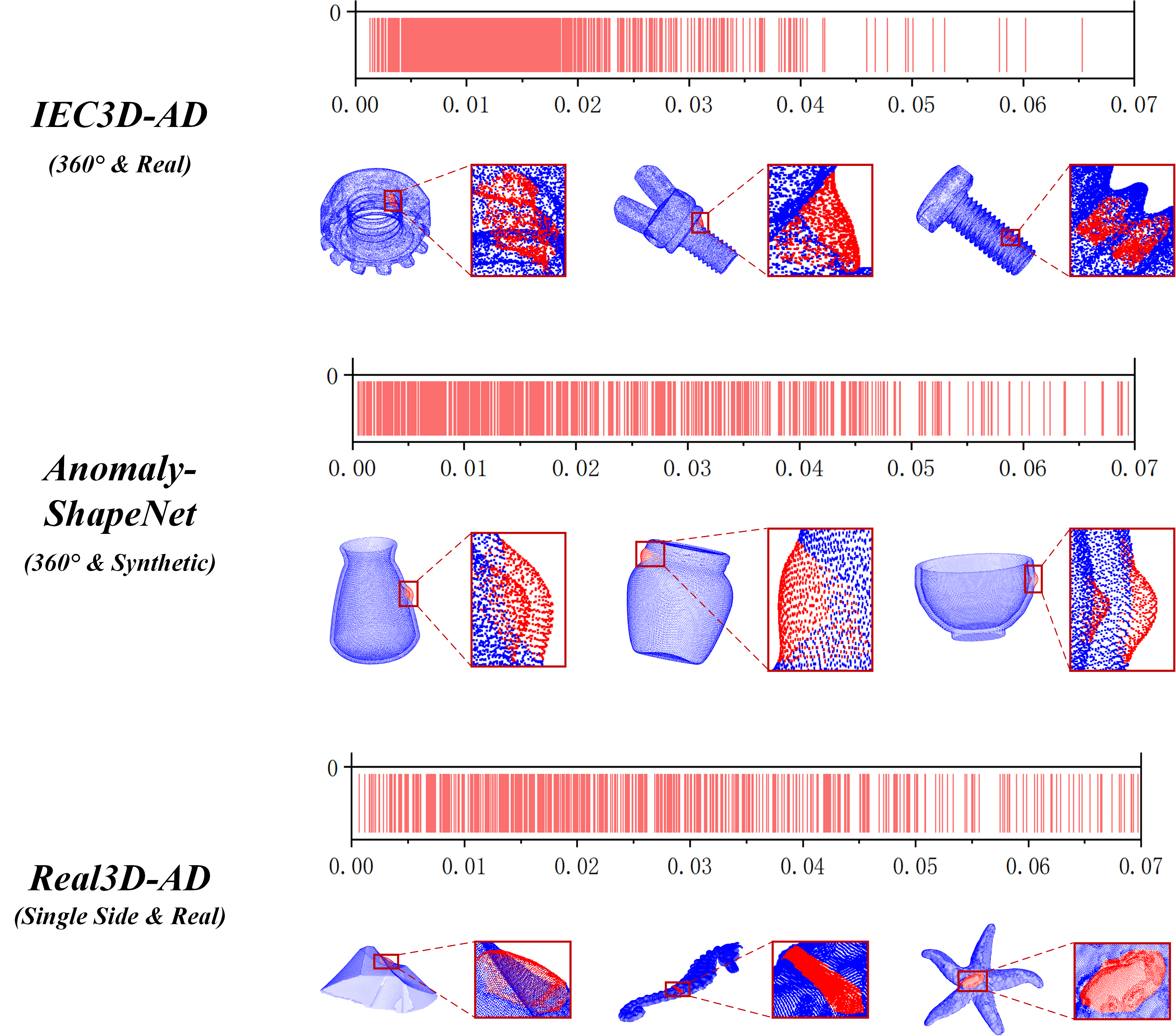}
    \caption{IEC3D-AD possesses a small defect ratio while ensuring data authenticity and spatial coverage. It can be seen that defects in real industrial scenarios are significantly different from others.}
    \label{fig0}
\end{figure}

\textbf{Limitations of existing datasets.} As of the completion of this paper, there are a total of four datasets on 3D industrial anomaly detection, such as MVTec 3D-AD \cite{DBLP:conf/visapp/BergmannJSS22}, Eyescandies \cite{DBLP:conf/accv/BonfiglioliTSFG22}, Real3D-AD \cite{DBLP:conf/nips/LiuXCLWLWZ23}, and Anomaly-ShapeNet \cite{DBLP:journals/corr/abs-2311-14897}. The above datasets have extensively promoted the development of this field and made significant progress. However, there are still two limitations: 1) \textit{Lack of samples in real industrial scenarios.} Data derived from actual industrial scenarios possess unique appearance characteristics that data from other fields cannot replace. The existing datasets contain a small portion of data from real industrial scenarios, with most of the data in MVTec 3D-AD and Real3D-AD coming from molds and toys, while the data in Eyescandies and Anomaly-ShapeNet are virtually synthesized. These samples have significant differences in data representation compared to complex industrial production environments, which leads to performance degradation of the algorithm in practical applications. 2) \textit{Difficult to simultaneously balance spatial coverage and data density.} It is reasonable to anticipate that spatial coverage and data density in point cloud datasets would be positively correlated, as comprehensive coverage naturally guarantees dense data points. Nevertheless, this is not the case with the current datasets, as a result of the inherent challenges in data acquisition environments and sources. For example, MVTec 3D-AD \cite{DBLP:conf/visapp/BergmannJSS22} and Real3D-AD \cite{DBLP:conf/nips/LiuXCLWLWZ23}, which are primarily composed of real-world data, frequently generate single-sided point clouds that do not accurately represent the complete geometry of objects. Consequently, they are incapable of detecting anomalies in their entirety and may overlook subtle defects that could be identified with a more comprehensive context. In contrast, whereas datasets such as Anomaly-ShapeNet \cite{DBLP:journals/corr/abs-2311-14897} provide substantial spatial coverage, they exhibit lower point cloud densities, with maximum resolutions limited to approximately 30K points. The low density is insufficient for facilitating high-resolution 3D inspections necessary for anomaly detection.

\textbf{Improvement of IEC3D-AD dataset.} In response to the above limitations, we construct IEC3D-AD dataset, where all samples are derived from real industrial production lines, and the collected point clouds achieve 360-degree full coverage. Unlike the molds and toys used in previous datasets, the authenticity of IEC3D-AD data fills the gap in the industrial 3D-AD field. At the same time, the proportion of defects has reached a leading level in the 3D-AD dataset. There are 160 samples in each category of IEC3D-AD, including a total of 100 normal samples and 60 abnormal samples. Most samples remain stable at around 200K, while only a few small molds are around 50K. In addition, IEC3D-AD still introduces two challenges that were not present in the previous dataset: 1) \textit{The samples in IEC3D-AD contain a large number of functional topological structures.} The structural design of industrial equipment components strictly follows physical functions, such as lead angle of thread or gear involute, and the defect characterization has more physical semantic information. 2) \textit{There are microscopic geometric distortions on the surface of real parts in IEC3D-AD.} Machining textures, such as turning tool marks and grinding lines, are present on the industrial equipment components surface, resulting in more microscopic geometric distortion. In comparison to other datasets, the properties of IEC3D-AD effectively represent the critical but challenging scenario of detecting minor defects in real industry scenarios. We demonstrate the above improvements and challenges in Fig. \ref{fig0}.

\textbf{Benchmark \& Novel Method.} To demonstrate the effectiveness of IEC3D-AD, we assess its performance in 3D-AD tasks via a thorough benchmark. This enables researchers to analyze performance discrepancies and possible advantages of various methodologies. What's more, we also introduce a novel unsupervised point cloud anomaly detection method (GMANet) inspired by generative 2D detection methods. Firstly, we generate synthetic point cloud generation (SPCG) based on geometric morphological analysis (GMA) to generate fake abnormal samples during the training process. Then, we extract fake abnormal feature distributions and their corresponding normal feature distributions in both expert and apprentice domains. We compare the differences between anomalous features and normal features using the prior anomalous regions of SPCG. Subsequently, we introduce a spatial discrepancy optimization to reduce the margin and enlarge the overlap between normal and abnormal point-level features. During testing, we input real abnormal samples to calculate abnormal point scores.

\textbf{Contributions.} In summary, we make three main contributions: 1) We introduce a novel industrial dataset of industrial equipment components for 3D-AD task. IEC3D-AD has excellent performance in terms of sample size and accuracy, and all of its samples come from actual industrial scenarios. 2) We propose a comprehensive benchmark for unsupervised 3D-AD task, which analyzes the effectiveness of the dataset. We believe that the IEC3D-AD dataset will fill the data gap of industrial equipment components. 3) We introduce a novel unsupervised 3D anomaly detection method based on high-quality synthetic point cloud samples. Numerous experiments indicate that our method achieves excellent performance.

\section{Related Work}
\label{sec:related}

\subsection{3D-AD Datasets} 
The maturity of 3D acquisition technology has solved the shortcomings of 2D data in-depth information and the relative position of objects. Recently, there are numerous research towards to 3D-AD datasets. MVTec Software proposed the first 3D-AD dataset named MVTec 3D-AD \cite{DBLP:conf/visapp/BergmannJSS22}. It contains more than 4000 high-resolution samples obtained from Zivid One Plus. Each of the ten different object categories contains a set of defect-free training and validation samples, as well as a set of samples with various defects. Eyecan-ai published a synthetic dataset called Eyecandies \cite{DBLP:conf/accv/BonfiglioliTSFG22}. This dataset provides multiple data modalities, such as RGB images, depth maps, and camera pose matrices. The ground truth includes geometric parameters, shadow parameters, and anomaly labels. Eyecandies and MVTec 3D-AD are both RGBD datasets that contain only single-view information. In order to solve this issue, Real3D-AD \cite{DBLP:conf/nips/LiuXCLWLWZ23} attempts to construct a novel 3D dataset that covers $360^{\circ}$ spatial information of different objects. This dataset contains 12 categories of objects, with 100 samples in each category. Meanwhile, the point cloud number of each sample ranges from 35K to 780K, significantly surpassing previous datasets. Recently, another excellent 3D dataset, Anomaly-ShapeNet \cite{10658462}, is comprised of 1600 point cloud samples organized into 40 distinct categories, offering a diverse and extensive dataset. The existing data lack actual industrial samples, which is also limited by the accuracy and quantity of point clouds.

\subsection{3D-AD Methods} 
In contrast to 2D-AD methods, the field of 3D anomaly detection continues to exhibit comparatively slower progress due to the inherent obstacles of the dataset. Bergmann \cite{10030323} and Rudolph \cite{10030168} both present a novel teacher-student network to detect geometric anomalies in 3D point clouds. A novel approach called BTF \cite{Horwitz_2023_CVPR} is proposed, which combines manually designed 3D representations using Fast Point Feature Histograms (FPFH) with a deep learning method based on color called PatchCore \cite{9879738}.  Chu \cite{pmlr-v202-chu23b} first uses neural implicit functions of signed distance fields to detect anomalies at the fine-grained per-point level, which effectively combines the complementary aspects of color and geometry. Kruse \cite{kruseSplatPose} proposes the novel 3D Gaussian splatting-based framework using Neural Radiance Fields (NeRFs), which precisely predicts the position of unseen perspectives in a way that can be differentiated and identifies any abnormalities in them. In order to better align point cloud structural information and semantic information, many methods \cite{liang2025taminganomaliesdownupsampling,cheng2025mc3dadunifiedgeometryawarereconstruction,liang2025look} have begun to explore the correlation between local point clouds and global point clouds. Several methods, such as CPMF \cite{cao2023complementary}, M3DM \cite{wang2023multimodal}, and MulSen-AD \cite{li2024multisensorobjectanomalydetection}, attempt utilizing multimodal fusion to address the issue of insufficient accuracy caused by using single data. Moreover, many mamba-based methods \cite{zhang2024point,han2024mamba3d,liu2024point,liang2024pointmamba,Wang2024PoinTrambaAH} using the state space model (SSM) have emerged, which achieve both linear complexity and long-range context learning abilities in point cloud data. The existing methods have shown excellent performance in custom point cloud data but still need to be validated in industrial fields, especially in practical production scenarios.

\section{IEC3D-AD Dataset}
\label{sec:data}

\subsection{Data source} 
\begin{figure}
    \centering
    \includegraphics[width=\linewidth]{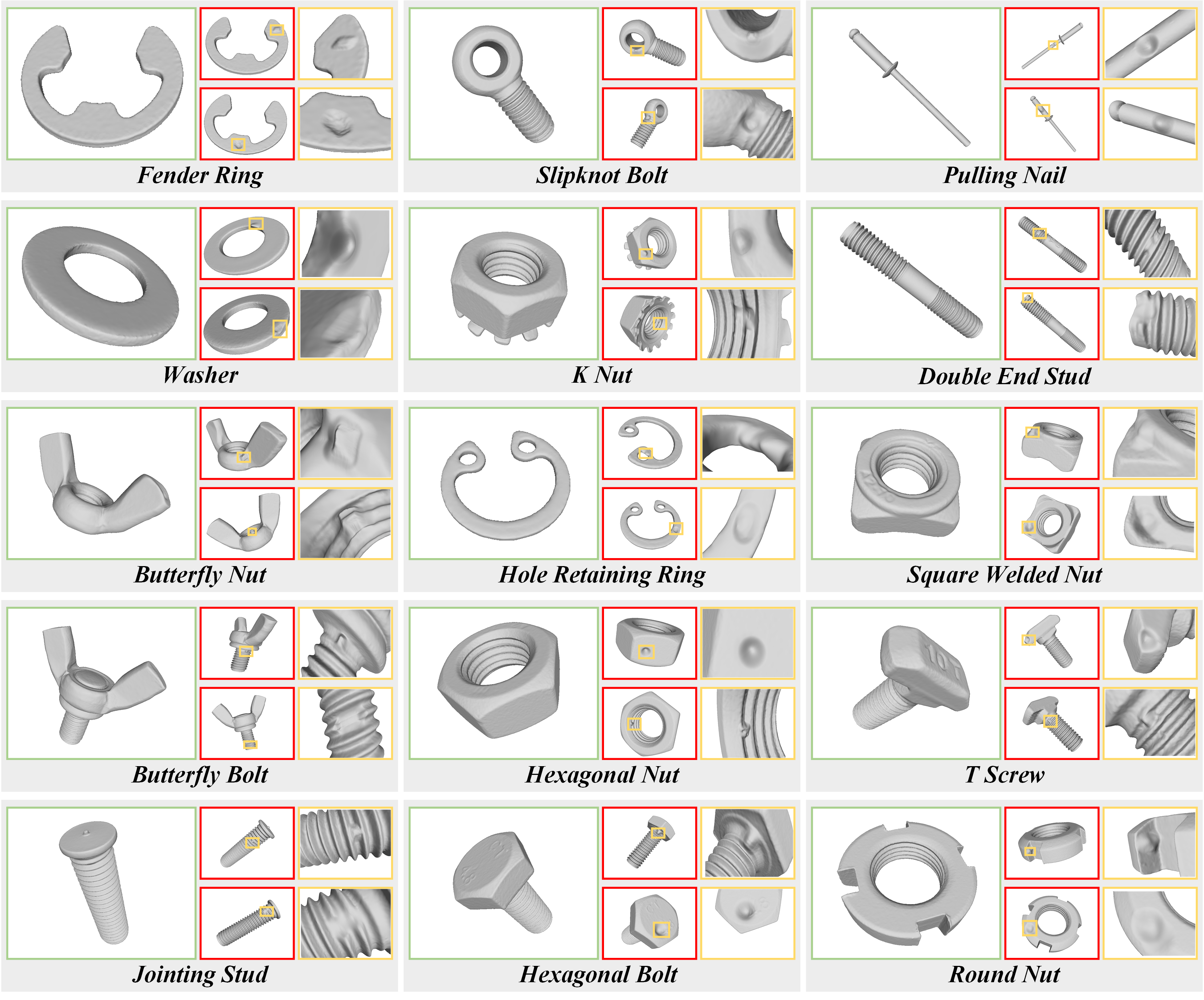}
    \caption{The example of IEC3D-AD for each category.}
    \label{fig1}
\end{figure}

In order to ensure the reliability of data collection, all samples in the IEC3D-AD dataset are obtained solely from the industrial processing production line. As shown in Fig. \ref{fig1}, the dataset encompasses 15 types of industrial equipment components, each type comprising normal and abnormal samples. The abnormal samples cover a total of 5 common processing defects, such as convex, concave, scratch, scar, and deformation. Each type of defect is caused by process parameters or tool quality issues during the machining process, which can affect the mechanical performance of the component.

\subsection{Collection equipment} 
To obtain accurate point cloud data, we conduct an acquisition system including two CCD cameras, a structured light, and a $360^{\circ}$ automatic turntable, as shown in Tab. \ref{tab:my_label}. The combination of two CCD cameras is a key component of this device, providing high-resolution and high-sensitivity imaging capabilities. The structure light projector emits distinct light patterns that create distinctive designs on the components, which are subsequently caught by the camera. The object is placed on the rotating turntable, and the rotation triggers the projection and photographing processes, allowing for precise measurements to be obtained from various perspectives. 

\begin{table}[htp]
    \centering
    \caption{The details of acquisition system.}
    \resizebox{\linewidth}{!}{
    \begin{tabular}{c|c|c|c}
    \toprule
    \toprule
    Serial Number & Device & Type  & Important Parameters \\
    \midrule
    \rowcolor[rgb]{ .906,  .902,  .902} 1     & Structure Light Projector &  DLP 3010    &  Display resolution $1280 \times 720$ \\ 
    2     & CCD Camera & Daheng MER-502-79U3M & 2/3" IMX250 Global shutter CMOS \\
    \rowcolor[rgb]{ .906,  .902,  .902} 3     & Triangle Bracket & Customization & - \\
    4     & Automatic Turntable & Customization & $360^{\circ}$ rotation \\ 
    \bottomrule
    \bottomrule
    \end{tabular}}%
    \label{tab:my_label}
\end{table}

\begin{figure}
    \centering
    \includegraphics[width=\linewidth]{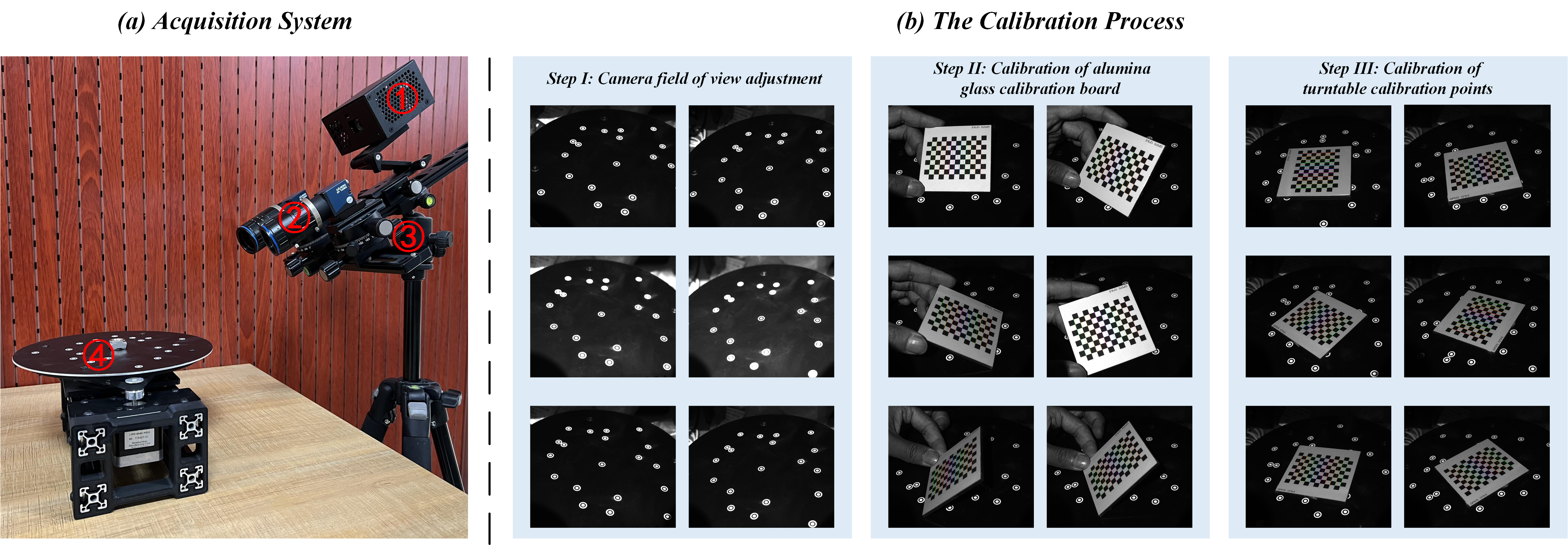}
    \caption{(a) Acquisition system. (b) The pipeline of calibration process. The relevant parameters generated through the calibration process can help us complete the coarse and fine registration tasks of point cloud data.}
    \label{figs1}
\end{figure}

In order to ensure that the system can provide accurate and reliable measurement results, we have constructed a detailed calibration process. Initially, the system conducts a camera calibration. The center-view fields of the two cameras, left and right, are adjusted to focus on the same point. After this, the level of light allowed in and the focus of each camera is adjusted to obtain a clear view in the middle of the field. Subsequently, the system proceeds with a calibration using a 5mm alumina calibration plate. The plate is flat on the revolving stage, and the calibration images are captured. The calibration plate will be moved several times to ensure the calibration is conducted within the entire camera view field. Following this, random incline angles and orientations are selected, capturing several calibration images. Consequently, a specific calibration algorithm \cite{zhang2000flexible} is executed. Prior to the initiation of 3D reconstruction, the calibration plate is placed at the central view field. Finally, the revolving stage spins at specified angles and captures calibration images. Based on these images, specific calibration parameters files are generated.  The calibration process is shown in Fig. \ref{figs1} (b).

\begin{table*}[htbp]
  \centering
  \caption{The statistics of IEC3D-AD dataset.}
  \resizebox{\linewidth}{!}{
    \begin{tabular}{c|c|ccc|c|ccc|c|c}
    \toprule
    \toprule
    \multirow{3}[6]{*}{Category} & \multicolumn{4}{c|}{Normal}   & \multicolumn{4}{c|}{Abnormal} & \multirow{3}[6]{*}{Total} & \multirow{3}[6]{*}{Anomaly Point Ratio} \\
\cmidrule{2-9}          & \multirow{2}[4]{*}{Sample Number} & \multicolumn{3}{c|}{Point Number (K)} & \multirow{2}[4]{*}{Sample Number} & \multicolumn{3}{c|}{Point Number (K)} &       &  \\
\cmidrule{3-5}\cmidrule{7-9}          &       & Avg.  & Max.  & Min.  &       & Avg.  & Max.  & Min.  &       &  \\
    \midrule
    Fender Ring & 100   & 29.21 & 45.01 & 22.52 & 60.00 & 29.94 & 37.65 & 26.39 & 160   & 1.18\% \\
    \rowcolor[rgb]{ .906,  .902,  .902} Washer & 100   & 56.63 & 87.11 & 43.56 & 60.00 & 58.08 & 71.00 & 47.60 & 160   & 0.96\% \\
    Butterfly Nut & 100   & 78.94 & 121.96 & 60.98 & 60.00 & 80.86 & 100.08 & 69.66 & 160   & 1.59\% \\
    \rowcolor[rgb]{ .906,  .902,  .902} Butterfly Bolt & 100   & 161.11 & 259.56 & 120.30 & 60.00 & 166.69 & 259.61 & 129.13 & 160   & 1.33\% \\
    Jointing Stud & 100   & 291.59 & 487.19 & 243.60 & 60.00 & 291.35 & 311.63 & 264.98 & 160   & 1.22\% \\
    \rowcolor[rgb]{ .906,  .902,  .902} Slipknot Bolt & 100   & 190.90 & 334.58 & 131.16 & 60.00 & 187.62 & 260.50 & 131.16 & 160   & 1.76\% \\
    K Nut & 100   & 208.73 & 313.27 & 156.63 & 60.00 & 215.66 & 239.83 & 192.54 & 160   & 1.34\% \\
    \rowcolor[rgb]{ .906,  .902,  .902} Hole Retaining Ring & 100   & 26.34 & 41.33 & 20.67 & 60.00 & 26.86 & 30.22 & 24.67 & 160   & 2.28\% \\
    Hexagonal Nut & 100   & 218.01 & 347.22 & 176.61 & 60.00 & 221.24 & 233.25 & 208.11 & 160   & 1.48\% \\
    \rowcolor[rgb]{ .906,  .902,  .902} Hexagonal Bolt & 100   & 209.69 & 388.49 & 194.26 & 60.00 & 209.64 & 232.66 & 162.05 & 160   & 1.18\% \\
    Pulling Nut & 100   & 70.93 & 112.36 & 56.18 & 60.00 & 72.10 & 84.59 & 65.98 & 160   & 2.24\% \\
    \rowcolor[rgb]{ .906,  .902,  .902} Double End Stud & 100   & 217.27 & 503.72 & 64.18 & 60.00 & 188.95 & 322.47 & 64.18 & 160   & 0.78\% \\
    Square Welded Nut & 100   & 153.97 & 255.20 & 127.60 & 60.00 & 154.25 & 162.69 & 149.44 & 160   & 1.27\% \\
    \rowcolor[rgb]{ .906,  .902,  .902} T Screw & 100   & 219.80 & 389.73 & 194.86 & 60.00 & 215.11 & 231.48 & 196.93 & 160   & 1.03\% \\
    Round Nut & 100   & 276.62 & 355.61 & 227.80 & 60.00 & 277.71 & 293.70 & 256.97 & 160   & 1.02\% \\
    \midrule
    \rowcolor[rgb]{ .906,  .902,  .902} Mean  & 100   & 160.65 & 269.49 & 122.73 & 60.00 & 159.74 & 191.42 & 132.65 & 160   & 1.35\% \\
    \bottomrule
    \bottomrule
    \end{tabular}}%
  \label{tab2}%
\end{table*}%

\subsection{Data Processing} 

\textbf{Coarse Registration.} Initially, by following the calibration processes mentioned above, we have successfully obtained both the internal and external properties of the camera, along with the distortion coefficient. Subsequently, utilizing these calibration parameters, we construct a binocular vision model. More precisely, we employ the inverse projection technique to transform the points obtained from binocular vision into three-dimensional point cloud data. We establish a fixed perspective as the reference coordinate system to standardize the data collected from various viewpoints of the component. Subsequently, we transform the point cloud data from all other perspectives to align with this reference coordinate system. We employ congruent sets of 4 super points (S4PCS) \cite{aiger20084} to efficiently identify matching pairs with angles that fall within a specified range to meet the coarse registration requirement.

\textbf{Fine Registration.} Following the initial registration, the multi-view point cloud images are concatenated but cannot be fused into a smooth whole. We achieve more precise stitching by using the position of the circular reference points after rough registration. We then utilize the least squares method to compute a more precise transformation matrix for these corresponding points. This matrix helps minimize the discrepancy between the predicted and actual positions of the marker points in each viewpoint. By utilizing the aforementioned transformation matrix, we may enhance the precision of aligning all point cloud data within the global coordinate system. To enhance the precision of fine registration, we employ the Fast and Robust Iterative Closest Point (FRICP) technique \cite{zhang2021fast} to seek the ideal alignment between two point clouds, not solely on the marked points. This process continues until the registration outcome satisfies the predetermined accuracy criteria.

\textbf{Denoising.} In order to minimize incorrect matching points as much as possible, we further denoise the point cloud data. Specifically, we calculate the density of neighboring points for each point and remove those points that deviate significantly from the mean, which can eliminate outliers. This process can be repeated until the predetermined threshold is met. Following denoising, we perform resampling procedures using the octree technique. This approach aims to retain the original shape of the point cloud while reducing the quantity of data points and simplifying further processing.

\textbf{Annotation.} During the annotation stage, we employ the CloudCompare software to precisely mark the areas of defects on the industrial equipment components at a granular level. This annotation procedure captures the precise form and dimensions of defects and their exact locations on the entire component, offering abundant defect data.

\subsection{Data statistics} 
As shown in Tab. \ref{tab2}, IEC3D-AD contains 100 normal and 60 abnormal samples in each category. We present in detail the average number of points, maximum number of points, minimum number of points, and anomaly point ratio for normal and abnormal samples under each category. It is obvious that IEC3D-AD has high point cloud quality. Unlike previous datasets, the proportion of abnormal samples in IEC3D-AD ranges from 0.96\% to 2.28\%, which poses more significant challenges in the construction of detection algorithms.

\begin{table}[htbp]
  \centering
  \caption{Comparison with other 3D dataset.}
  \resizebox{\linewidth}{!}{
    \begin{tabular}{c|c|c|c|c|c|c|c}
    \toprule
    \toprule
    Dataset & Categories & Number & Abnormal Class & Min. Point & Max. Point & Total Point & Type \\
    \midrule
    \rowcolor[rgb]{ .906,  .902,  .902} MVTecAD-3D & 10    & 3604  & 5     & 10K   & 20K   & -     & Real \\
    Eyecandies & 10    & 15500 & 3     & -     & -     & -     & Syn \\
    \rowcolor[rgb]{ .906,  .902,  .902} Real3D-AD & 12    & 1200  & 2     & 35K   & 780K  & 224,720K & Real \\
    Anomaly-ShapeNet & 40    & 1600  & 6     & 8K    & 30K   & -     & Syn \\
    \midrule
    \rowcolor[rgb]{ .906,  .902,  .902} IECAD-3D & 15    & 2400  & 5     & 14K   & 503K  & 340,335K & Real \\
    \bottomrule
    \bottomrule
    \end{tabular}}%
  \label{tab3}%
\end{table}%


\begin{figure}
    \centering
    \includegraphics[width=\linewidth]{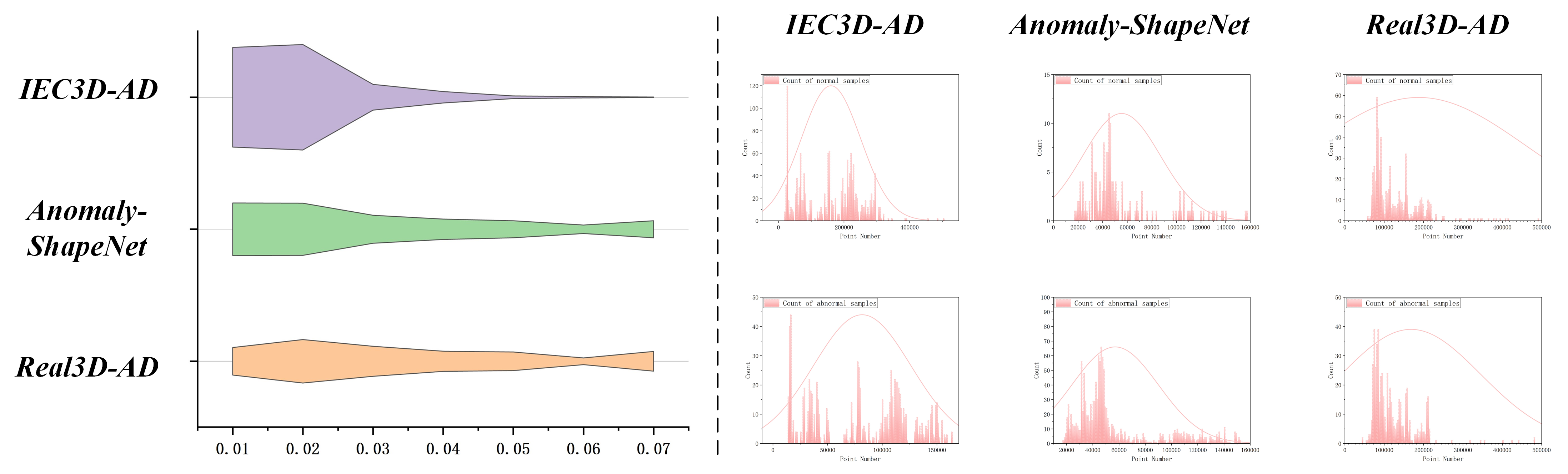}
    \caption{Comparison of abnormal proportions distribution and point cloud number of different datasets.}
    \label{figc1}
\end{figure}

\subsection{Comparison with other 3D-AD datasets}
1) \textit{Overall comparison:} As shown in Tab. \ref{tab3}, in all real scenarios, IECAD-3D has the highest number of categories, and its total sample number is only lower than MVTecAD-3D. Moreover, the total number of point clouds of IECAD-3D is the highest that others, which has almost 1.5 times compared to the current largest dataset Real3D-AD. Compared to datasets with synthetic samples, the data size of IEC3D-AD cannot match the sample size generated by software, as it is challenging to generate abnormal samples in real industrial scenarios. 2) \textit{Detail comparison:} As shown in Fig. \ref{fig0}, we visualize the details of different datasets. It can be seen that the actual location of defects in our dataset is very random and may be inside complex component structures. The defects of the other two datasets are only on the surface. What's more, the overall appearance of Anomaly-ShapeNet is relatively simple, and the test sample of Real3D-AD is single-sided. Our dataset simultaneously ensures both spatial coverage and sample authenticity. 3) \textit{Comparison of defect proportion:} As shown in Fig. \ref{figc1}, we display defect proportion of different datasets. Compared to other datasets, our data mostly falls within the range of 0-0.02 proportions, which can better achieve the task of detecting extremely tiny defects. We also visualize the distribution of point cloud numbers, and we find that their actual values are inconsistent with the number of tables in the original paper. 

\section{Methodology}
\label{sec:method}

\subsection{Overview} 
We define IEC3D-AD dataset as $\mathbb{D}$, which consists of normal samples and abnormal samples with a rate $\lambda \in (0,1)$, e.g. $\mathbb{D}=\lambda{\mathbb{D}}^{+}+(1-\lambda){\mathbb{D}}^{-}$. Specifically, the normal samples are divided into ${\mathbb{D}}^{+}_{train} = \left \{ d^+_{train,i} \right \}_{i=1}^n$ and ${\mathbb{D}}^{+}_{test} = \left \{d^+_{test,j} \right \}_{j=1}^m$, these parts are disjoint from each other, ${\mathbb{D}}^{+}_{train} \cap {\mathbb{D}}^{+}_{test} = \emptyset$. Like most unsupervised methods, we utilizes ${\mathbb{D}}^{+}_{train}$ as the training data. The testing data is consist of ${\mathbb{D}}^{+}_{test}$ and ${\mathbb{D}}^{-} =  \left \{ d^{-}_{k} \right \}_{k=1}^l $. This data partitioning method helps to quickly conduct unsupervised algorithms, which can identify normal and abnormal samples.

\begin{figure*}
    \centering
    \includegraphics[width=\linewidth]{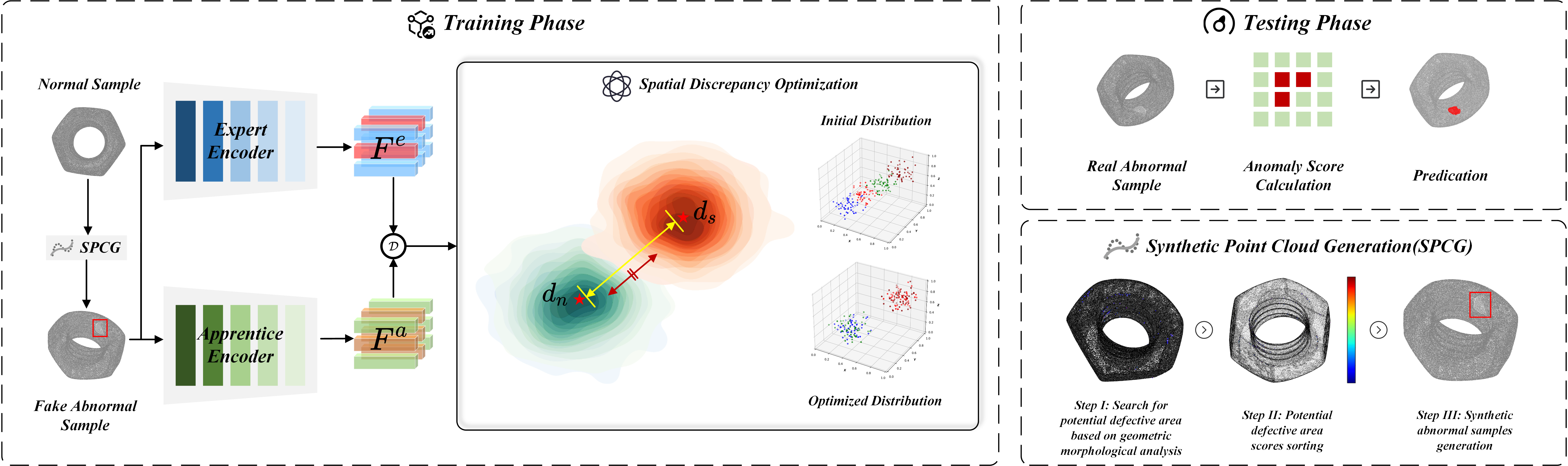}
    \caption{The pipeline of GMANet. During the training phase, we conduct a novel synthetic point cloud generation (SPCG) module based on geometric morphological analysis (GMA) to generate synthetic abnormal samples. Then, our spatial discrepancy optimization (SDO) module simultaneously reduces the margin between the normal and synthetic abnormal feature distribution extracted from different encoders and expands the overlap of the feature distribution of similar features between different encoders. During testing phase, we conduct an anomaly score calculation function to detect abnormal points.}
    \label{fig4}
\end{figure*}

The overview of our method is shown in Fig. \ref{fig4}. Above, we design a synthetic point cloud generation (SPCG) module based on geometric morphological analysis (GMA) to generate synthetic abnormal samples $\mathbb{D}_{syn}$, e.g. $\mathbb{D}_{syn}=\mathcal{G}({\mathbb{D}}^{+}_{train})$. Then, we extract the normal features $\mathit{F}_{n}=\left \{ \mathit{F}_{n}^{a}+\mathit{F}_{n}^{e}\right \}$ and abnormal features $\mathit{F}_a=\left \{ \mathit{F}_{a}^{a}+\mathit{F}_{a}^{e}\right \}$ corresponding to respective point clouds in the expert domains and apprentice domains. Subsequently, we design a spatial discrepancy optimization (SDO) module to simultaneously reduce the margin between the normal and synthetic abnormal feature distribution extracted from different encoders, and expand the overlap of the feature distribution of similar features between expert and apprentice encoders. Finally, we calculate the anomaly score for individual points via a function $\mathcal{A}: d \rightarrow \mathbb{R}$.

\begin{algorithm}[h]
\renewcommand{\algorithmicrequire}{\textbf{Input:}}
\renewcommand{\algorithmicensure}{\textbf{Output:}}
\caption{SPGC Based on GMA}
\label{alg1}
\begin{algorithmic}[1]
\REQUIRE Normal sample $\mathbb{D}^{+}_{\text{train}} = \left \{ d^+_{\text{train},i} \right \}_{i=1}^n$; Perturbation strengths $\mathbb{E} =  \left \{ \epsilon_p \right \}_{i=1}^n$, where $0 \le \epsilon_p \le 1$; Perturbation modes $\mathbb{M} = \left \{ \varepsilon_p \right \}_{i=1}^n$, where $\varepsilon_p =$ \textsc{True} $\OR$ \textsc{False}
\ENSURE Synthetic abnormal samples $\mathbb{D}_{\text{syn}} =  \left \{ d_{\text{syn},i} \right \}_{i=1}^n$
\FOR{${d^+_{\text{train},i}},{\epsilon_p},{\varepsilon_p}$ in ${\mathbb{D}^{+}_{\text{train}}}, {\mathbb{E}}, {\mathbb{M}}$}
\STATE $N \leftarrow |d^+_{\text{train},i}|$ 
\STATE $M \leftarrow \text{int}(N \times 0.015)$ 
\STATE $\mathcal{CEN} \leftarrow \textsc{Mean}(d^+_{\text{train},i})$ 
\STATE $\mathcal{CLS} \leftarrow \textsc{KD-Tree Clustering}(d^+_{\text{train},i})$ 
    \FOR {$cls_j$ in $\mathcal{CLS}$}
    \STATE $cen_j \leftarrow \textsc{Mean}(cls_j)$
    \STATE $v_j, t_j \leftarrow \textsc{Poisson Reconstruction}(cls_j)$ 
    \STATE $cur_j \leftarrow \textsc{Curvature Calculation}(v_j, t_j)$ 
    \IF{$cur_j > 0.2$} 
        \STATE $sco_j \leftarrow \|cen_j - \mathcal{CEN}\| + cur_j$
    \ENDIF
    \ENDFOR
\STATE $\mathbf{I} \leftarrow \text{argmax}(\mathcal{SCO} = \left \{ sco_j \right \}_{j=1}^m )$ 
\IF{$\varepsilon_p$}
\STATE $d_{\text{syn},i} \leftarrow$ \textsc{AddPoints}($d^+_{\text{train},i},\epsilon_p$)
\ELSE
\STATE $d_{\text{syn},i} \leftarrow$ \textsc{RemovePoints}($d^+_{\text{train},i},\epsilon_p$)
\ENDIF
\ENDFOR
\RETURN $d_{\text{syn},i}$
\end{algorithmic}
\end{algorithm}

\subsection{Synthetic Point Cloud Generation} 
Inspired by previous 2D unsupervised anomaly detection methods \cite{9681338,9000806,9212099}, synthetic data generation through adversarial learning, GAN, or VAE generates samples that are comparable to actual data. For point clouds, the complexity and dimensionality of data make it challenging to produce effective synthetic abnormal samples. In order to resolve the aforementioned issue, we developed the synthetic point cloud generation (SPCG) module based on geometric morphological analysis. This module conducts curvature analysis on the point cloud clusters of normal samples to produce more realistic abnormal samples.

Specifically, we firstly use KD-Tree to cluster the point cloud of normal sample, the clusters can be represents as $\mathcal{CLS} = \left \{ cls_j \right \}_{j=1}^n$. For each cluster $cls_j$, we utilize poisson reconstruction to calculate the curvatures. If the curvature exceeds our established threshold of 0.2, it is categorized as a potential defective area. Meanwhile, we considered the distance between the center point of the area and the center of the overall point cloud, and added it to the curvature as the final scoring metric $\mathcal{SCO} = \left \{ sco_j \right \}_{j=1}^m$. Then, we choose the area with the highest score as the generation area. We randomly set the boolean value based on the perturbation modes $\mathbb{M}$ to determine whether the overall appearance of the defect require adding or removing points. Finally, we add or remove point clouds to the original point cloud based on the total number $N$ and perturbation strengths $\mathbb{E}$. The entire processing process is shown in Alg. \ref{alg1}. 


\subsection{Spatial Discrepancy Optimization} 
Due to the inability to introduce defective samples during the training phase, most methods involve reconstructing the point cloud distribution \cite{10658462} or relying on memory banks \cite{DBLP:conf/nips/LiuXCLWLWZ23} for anomaly score calculation. Thanks to the introduction of SPCG, we are able to effectively compare the differences between abnormal and normal features during the training process. Our intuition is that we can optimize the difference distribution to reduce the margin and enlarge the overlap in feature space.

Above all, we utilize backbone of PointNet++ \cite{DBLP:conf/nips/QiYSG17} as our expert and apprentice encoder which are common used in other SOTA methods. The features between the expert and apprentice domains are formulated as:

\begin{equation}
\label{eq1}
F^e = \Phi_e(d_{\text{syn},i};\theta_e), \\
F^a = \Phi_a(d_{\text{syn},i};\theta_a)
\end{equation}

where $d_{\text{syn},i}$ denotes the synthetic abnormal samples; $\Phi_e$ and $\theta_e$ represent the expert encoder and parameters; $\Phi_a$ and $\theta_a$ mean the apprentice encoder and parameters, and $\theta_e$ remains frozen while $\theta_a$ is tuned during training phase. 

Subsequently, we utilize the coordinate of the synthetic abnormal area as mask to split normal features and abnormal features, and calculates the point-level mean square errors as feature distribution difference:

\begin{equation}
\label{eq2}
d_{n,i} = \parallel F_n^a, F_n^e \parallel_2, \\
d_{s,i} = \parallel F_a^a, F_a^e \parallel_2
\end{equation}

Moreover, the ratio of individual difference to the average serves as a significant indicator of point-level importance during the training phase. Thus, we refer focal loss \cite{8237586} to design the weight for each difference, which can be formulated as:

\begin{equation}
\label{eq3}
\omega_{n,i} = [ \frac{d_{n,i}\times N_n}{ {\textstyle \sum_{i=1}^{N_n}}d_{p,i} }]^\alpha,
\omega_{s,i} = [ \frac{d_{s,i}\times N_s}{ {\textstyle \sum_{i=1}^{N_s}}d_{p,i} }]^{-\alpha}
\end{equation}

where $0 \le \alpha$ is a parameter to adjust $\omega$, and the exponent for normal and abnormal samples are opposite. Thus, normal points exhibiting a difference greater than the average are amplified, while the weights of anomalous points with a difference less than the average are augmented. Based on the above parameters, we can simultaneously minimize $d_n$ and maximize $d_s$, which can effectively transfer expert domain knowledge to the apprentice domain:

\begin{equation}
\label{eq4}
\mathcal{L} = \frac{ {\textstyle \sum_{i=1}^{N_n}{\omega_{n,i}} \times {d_{n,i}} - \sum_{j=1}^{N_s}{\omega_{s,j}} \times d_{s,j}} }{ {\textstyle \sum_{i=1}^{N_n}}{\omega_{n,i}}+{\textstyle \sum_{j=1}^{N_s}}{\omega_{s,j}}} 
\end{equation}

The process of $Eq.$ \ref{eq4} is shown by the yellow and red bidirectional arrows in Fig. \ref{fig4}, where the yellow bidirectional arrow represents reducing the margin between normal and abnormal features in the expert domain and the apprentice domain, and the red bidirectional arrow represents enlarging the overlap of abnormal and normal areas further apart in feature distribution.

\begin{table*}[htbp]
  \centering
  \caption{AUROC and AUPR results under object level. The bold represents the best results.}
  \resizebox{\linewidth}{!}{
    \begin{tabular}{c|cc|cc|cc|cc|cc|cc|cc|cc|cc|cc|cc}
    \toprule
    \toprule
    \multirow{3}[6]{*}{Category} & \multicolumn{4}{c|}{BTF}      & \multicolumn{4}{c|}{M3DM}     & \multicolumn{6}{c|}{PatchCore}                & \multicolumn{2}{c|}{\multirow{2}[4]{*}{Reg3D-AD}} & \multicolumn{2}{c|}{\multirow{2}[4]{*}{IMRNet}} & \multicolumn{2}{c|}{\multirow{2}[4]{*}{R3D-AD}} & \multicolumn{2}{c}{\multirow{2}[4]{*}{GMANet (Ours)}} \\
\cmidrule{2-15}          & \multicolumn{2}{c|}{Raw} & \multicolumn{2}{c|}{FPFH} & \multicolumn{2}{c|}{PointMAE} & \multicolumn{2}{c|}{PointBERT} & \multicolumn{2}{c|}{FPFH} & \multicolumn{2}{c|}{FPFH+Raw} & \multicolumn{2}{c|}{PointMAE} & \multicolumn{2}{c|}{} & \multicolumn{2}{c|}{} & \multicolumn{2}{c|}{} & \multicolumn{2}{c}{} \\
\cmidrule{2-23}          & AUROC & AUPR & AUROC & AUPR & AUROC & AUPR & AUROC & AUPR & AUROC & AUPR & AUROC & AUPR & AUROC & AUPR & AUROC & AUPR & AUROC & AUPR & AUROC & AUPR & AUROC & AUPR \\
    \midrule
    \rowcolor[rgb]{ .851,  .851,  .851} Fender Ring & 0.6490  & 0.6850  & 0.3950  & 0.4260  & 0.5920  & 0.5910  & 0.5070  & 0.5550  & 0.9508  & 0.9017  & 0.9289  & 0.8955  & 0.6875  & 0.6823  & 0.6414  & 0.6077  & 0.6201  & 0.5965  & 0.6688  & 0.5968  & \textbf{0.9633 } & \textbf{0.9702 } \\
    Washer & 0.6570  & 0.6950  & 0.4740  & 0.4770  & 0.8270  & 0.8260  & 0.7780  & 0.7660  & 0.9411  & 0.9226  & 0.9031  & 0.8829  & 0.7203  & 0.7058  & 0.8033  & 0.8004  & 0.7995  & 0.8121  & 0.8391  & 0.7597  & \textbf{0.9511 } & \textbf{0.9558 } \\
    \rowcolor[rgb]{ .851,  .851,  .851} Butterfly Nut & 0.6650  & 0.6670  & 0.6100  & 0.7170  & 0.6640  & 0.6020  & 0.6140  & 0.5610  & 0.8250  & 0.8078  & 0.7236  & 0.7478  & 0.6869  & 0.6785  & 0.6731  & 0.6644  & 0.6856  & 0.6444  & 0.6668  & 0.6986  & \textbf{0.8556 } & \textbf{0.8181 } \\
    Butterfly Bolt & 0.6920  & 0.6760  & 0.6280  & 0.6480  & 0.4740  & 0.5790  & 0.5460  & 0.6060  & 0.8747  & 0.8223  & 0.8086  & 0.7844  & 0.7319  & 0.7639  & 0.8828  & \textbf{0.9111 } & 0.8984  & 0.9104  & \textbf{0.9053 } & 0.8972  & 0.8464  & 0.8028  \\
    \rowcolor[rgb]{ .851,  .851,  .851} Joingting Stud & 0.7520  & 0.7810  & 0.7600  & 0.7780  & 0.5870  & 0.6000  & 0.5640  & 0.5810  & 0.9133  & 0.9150  & 0.8908  & 0.8827  & 0.7681  & 0.7537  & 0.8056  & 0.7682  & 0.8372  & 0.7511  & 0.7612  & 0.7702  & \textbf{0.9367 } & \textbf{0.9404 } \\
    Slipknot Bolt & 0.7340  & 0.7320  & 0.3690  & 0.4250  & 0.4690  & 0.5200  & 0.5520  & 0.5850  & 0.9292  & 0.9587  & 0.9136  & 0.8715  & 0.6161  & 0.6349  & 0.6839  & 0.6777  & 0.6984  & 0.6964  & 0.6663  & 0.6971  & \textbf{0.9564 } & \textbf{0.9657 } \\
    \rowcolor[rgb]{ .851,  .851,  .851} K Nut & 0.6740  & 0.6410  & 0.3240  & 0.4000  & 0.5330  & 0.5640  & 0.6530  & 0.6550  & 0.7306  & 0.6913  & 0.8494  & 0.8443  & 0.7211  & 0.7175  & \textbf{0.8808 } & 0.9048  & 0.8529  & \textbf{0.9290 } & 0.8350  & 0.8951  & 0.7042  & 0.6760  \\
    Hole Retaining Ring & 0.6810  & 0.6650  & 0.4410  & 0.4690  & 0.7670  & 0.6930  & 0.7430  & 0.7310  & 0.9331  & 0.9001  & 0.9303  & 0.8831  & 0.6172  & 0.5530  & 0.6194  & 0.5386  & 0.6093  & 0.5523  & 0.6146  & 0.5296  & \textbf{0.9492 } & \textbf{0.9461 } \\
    \rowcolor[rgb]{ .851,  .851,  .851} Hexagonal Nut & 0.7790  & 0.7900  & 0.4530  & 0.4990  & 0.7040  & 0.6860  & 0.5310  & 0.5740  & 0.8569  & 0.8539  & \textbf{0.9097 } & \textbf{0.8986 } & 0.6758  & 0.6574  & 0.8253  & 0.8294  & 0.8368  & 0.7937  & 0.8476  & 0.7974  & 0.8739  & 0.8621  \\
    Hexagonal Bolt & 0.6800  & 0.6960  & 0.3970  & 0.4330  & 0.4970  & 0.4750  & 0.4470  & 0.5020  & 0.8475  & 0.8193  & 0.8597  & 0.7911  & 0.5272  & 0.5306  & 0.6550  & 0.6448  & 0.6222  & 0.6367  & 0.6465  & 0.6417  & \textbf{0.8925 } & \textbf{0.8238 } \\
    \rowcolor[rgb]{ .851,  .851,  .851} Pulling Nail & 0.6800  & 0.6260  & 0.6320  & 0.6210  & 0.7800  & 0.7950  & 0.7420  & 0.7180  & 0.8750  & 0.8577  & \textbf{0.9078 } & \textbf{0.9136 } & 0.6958  & 0.6803  & 0.8514  & 0.8754  & 0.8390  & 0.8672  & 0.8732  & 0.8662  & 0.8500  & 0.7996  \\
    Double End Sutd & 0.6230  & 0.6260  & 0.4880  & 0.5180  & 0.3720  & 0.4770  & 0.4580  & 0.5580  & 0.8067  & 0.7947  & 0.8400  & 0.7909  & 0.5431  & 0.6010  & 0.6956  & 0.7775  & 0.7110  & 0.7321  & 0.7248  & 0.7468  & \textbf{0.8503 } & \textbf{0.8174 } \\
    \rowcolor[rgb]{ .851,  .851,  .851} Square Welded Nut & 0.6810  & 0.6900  & 0.4200  & 0.4690  & 0.6580  & 0.6930  & 0.6560  & 0.6900  & 0.9006  & 0.9167  & 0.9078  & 0.8678  & 0.6675  & 0.6308  & 0.8456  & 0.8359  & 0.8366  & 0.8667  & 0.8372  & 0.8365  & \textbf{0.9239 } & \textbf{0.9210 } \\
    T Screw & 0.7600  & 0.7620  & 0.4280  & 0.4620  & 0.5020  & 0.5340  & 0.4700  & 0.5190  & 0.8244  & 0.8535  & \textbf{0.8717 } & \textbf{0.9304 } & 0.6206  & 0.6086  & 0.8161  & 0.8286  & 0.7855  & 0.8128  & 0.8671  & 0.8571  & 0.8311  & 0.8119  \\
    \rowcolor[rgb]{ .851,  .851,  .851} Round Nut & 0.6330  & 0.6460  & 0.3890  & 0.4360  & 0.5210  & 0.5240  & 0.4960  & 0.5010  & 0.8400  & 0.8116  & 0.8125  & 0.7865  & 0.4861  & 0.5188  & 0.5981  & 0.6023  & 0.6257  & 0.6348  & 0.6358  & 0.6481  & \textbf{0.8203 } & \textbf{0.8072 } \\
    \midrule
    Mean  & 0.6893  & 0.6919  & 0.4805  & 0.5185  & 0.5965  & 0.6106  & 0.5838  & 0.6068  & 0.8699  & 0.8551  & 0.8705  & 0.8514  & 0.6510  & 0.6478  & 0.7518  & 0.7511  & 0.7505  & 0.7491  & 0.7593  & 0.7492  & \textbf{0.8803 } & \textbf{0.8612 } \\
    \bottomrule
    \bottomrule
    \end{tabular}}%
  \label{tab4}%
\end{table*}%

\begin{table*}[htbp]
  \centering
  \caption{AUROC and AUPR results under point level. The bold represents the best results.}
  \resizebox{\linewidth}{!}{
    \begin{tabular}{c|cc|cc|cc|cc|cc|cc|cc|cc|cc|cc|cc}
    \toprule
    \toprule
    \multirow{3}[6]{*}{Category} & \multicolumn{4}{c|}{BTF}      & \multicolumn{4}{c|}{M3DM}     & \multicolumn{6}{c|}{PatchCore}                & \multicolumn{2}{c|}{\multirow{2}[4]{*}{Reg3D-AD}} & \multicolumn{2}{c|}{\multirow{2}[4]{*}{IMRNet}} & \multicolumn{2}{c|}{\multirow{2}[4]{*}{R3D-AD}} & \multicolumn{2}{c}{\multirow{2}[4]{*}{GMANet (Ours)}} \\
\cmidrule{2-15}          & \multicolumn{2}{c|}{Raw} & \multicolumn{2}{c|}{FPFH} & \multicolumn{2}{c|}{PointMAE} & \multicolumn{2}{c|}{PointBERT} & \multicolumn{2}{c|}{FPFH} & \multicolumn{2}{c|}{FPFH+Raw} & \multicolumn{2}{c|}{PointMAE} & \multicolumn{2}{c|}{} & \multicolumn{2}{c|}{} & \multicolumn{2}{c|}{} & \multicolumn{2}{c}{} \\
\cmidrule{2-23}          & AUROC & AUPR & AUROC & AUPR & AUROC & AUPR & AUROC & AUPR & AUROC & AUPR & AUROC & AUPR & AUROC & AUPR & AUROC & AUPR & AUROC & AUPR & AUROC & AUPR & AUROC & AUPR \\
    \midrule
    \rowcolor[rgb]{ .851,  .851,  .851} Fender Ring & 0.7660  & 0.0440  & 0.4910  & 0.0040  & 0.6300  & 0.0120  & 0.5370  & 0.0090  & 0.8295  & 0.1251  & 0.8268  & 0.0680  & 0.8052  & 0.0536  & 0.7873  & 0.0311  & 0.7491  & 0.0306  & 0.8046  & 0.0331  & \textbf{0.8386 } & \textbf{0.1397 } \\
    Washer & 0.7970  & 0.0850  & 0.5250  & 0.0040  & 0.6930  & 0.0160  & 0.6610  & 0.0110  & 0.8797  & 0.3036  & 0.8304  & 0.1221  & 0.7869  & 0.0335  & 0.8235  & 0.0783  & 0.8335  & 0.0802  & 0.8428  & 0.0771  & \textbf{0.8808 } & \textbf{0.2981 } \\
    \rowcolor[rgb]{ .851,  .851,  .851} Butterfly Nut & 0.7130  & 0.0190  & 0.4550  & 0.0050  & 0.5470  & 0.0080  & 0.5340  & 0.0070  & 0.6171  & 0.0152  & 0.6528  & 0.0138  & 0.7332  & 0.0463  & 0.7516  & 0.0381  & 0.7546  & 0.0370  & 0.7692  & 0.0369  & \textbf{0.7712 } & \textbf{0.0643 } \\
    Butterfly Bolt & 0.6640  & 0.0150  & 0.4360  & 0.0040  & 0.6720  & 0.0780  & 0.6440  & 0.0750  & 0.5864  & 0.0125  & 0.6732  & 0.0818  & 0.7865  & 0.2895  & 0.8193  & 0.3668  & \textbf{0.9385 } & \textbf{0.3776 } & 0.7824  & 0.3608  & 0.7797  & 0.3140  \\
    \rowcolor[rgb]{ .851,  .851,  .851} Joingting Stud & 0.6710  & 0.0340  & 0.6720  & 0.0130  & 0.3680  & 0.0040  & 0.3880  & 0.0040  & 0.5728  & 0.0321  & 0.5858  & \textbf{0.0406 } & 0.4563  & 0.0083  & 0.4758  & 0.0089  & 0.4972  & 0.0092  & 0.4542  & 0.0085  & \textbf{0.5932 } & 0.0360  \\
    Slipknot Bolt & 0.6280  & 0.0290  & 0.5460  & 0.0070  & 0.5100  & 0.0320  & 0.5080  & 0.0340  & 0.5787  & 0.0321  & 0.5760  & 0.0185  & \textbf{0.6481 } & \textbf{0.1017 } & 0.6164  & 0.0831  & 0.5960  & 0.0812  & 0.6116  & 0.0872  & 0.5906  & 0.0331  \\
    \rowcolor[rgb]{ .851,  .851,  .851} K Nut & 0.6140  & 0.0080  & 0.5280  & 0.0050  & 0.5550  & 0.0090  & 0.5220  & 0.0070  & 0.5722  & 0.0066  & 0.6516  & 0.0352  & 0.7370  & 0.0406  & 0.7886  & 0.0787  & 0.8156  & 0.0800  & 0.8121  & 0.0803  & \textbf{0.8273 } & \textbf{0.0968 } \\
    Hole Retaining Ring & 0.7560  & 0.0450  & 0.5320  & 0.0100  & 0.6370  & 0.0390  & 0.6370  & 0.0390  & 0.7239  & 0.0538  & 0.7109  & 0.0297  & 0.7817  & 0.0240  & 0.7686  & 0.0225  & 0.7503  & 0.0217  & 0.8060  & 0.0216  & \textbf{0.8160 } & \textbf{0.0557 } \\
    \rowcolor[rgb]{ .851,  .851,  .851} Hexagonal Nut & 0.6020  & 0.0130  & 0.5400  & 0.0120  & 0.5040  & 0.0060  & 0.4790  & 0.0050  & 0.5485  & 0.0100  & 0.6180  & 0.0460  & 0.6807  & 0.0265  & 0.6914  & 0.0592  & \textbf{0.7250 } & 0.0603  & 0.6625  & 0.0612  & 0.6551  & \textbf{0.0604 } \\
    Hexagonal Bolt & 0.6880  & 0.0240  & 0.5550  & 0.0050  & 0.4510  & 0.0040  & 0.4610  & 0.0040  & 0.6056  & 0.0112  & 0.6257  & 0.0375  & 0.5175  & 0.0090  & 0.5870  & 0.0161  & 0.6013  & 0.0153  & 0.5657  & \textbf{0.0168 } & \textbf{0.6182 } & 0.0114  \\
    \rowcolor[rgb]{ .851,  .851,  .851} Pulling Nail & 0.7260  & 0.0190  & 0.6140  & 0.0150  & 0.5830  & 0.0110  & 0.5710  & 0.0110  & 0.6458  & 0.0271  & 0.7009  & 0.1170  & 0.5607  & 0.0447  & 0.6459  & 0.1314  & 0.6773  & \textbf{0.1379 } & 0.6211  & 0.1348  & \textbf{0.6504 } & 0.1272  \\
    Double End Sutd & 0.7850  & 0.0310  & 0.5180  & 0.0030  & 0.6030  & 0.0330  & 0.6130  & 0.0260  & 0.6250  & 0.0128  & 0.6328  & 0.0585  & 0.5681  & 0.0220  & 0.6213  & 0.1010  & 0.6312  & 0.1030  & 0.6062  & 0.0980  & \textbf{0.6470 } & \textbf{0.1121 } \\
    \rowcolor[rgb]{ .851,  .851,  .851} Square Welded Nut & 0.7300  & 0.0380  & 0.5580  & 0.0050  & 0.5710  & 0.0080  & 0.5570  & 0.0080  & 0.6449  & 0.0206  & 0.6948  & 0.1065  & 0.6774  & 0.0183  & 0.7107  & 0.0588  & \textbf{0.7295 } & 0.0610  & 0.6876  & 0.0594  & 0.7245  & \textbf{0.0627 } \\
    T Screw & 0.6750  & 0.0080  & 0.5120  & 0.0060  & 0.4760  & 0.0050  & 0.4790  & 0.0050  & 0.5772  & 0.0055  & 0.6300  & 0.0893  & 0.6401  & 0.0211  & 0.7019  & 0.0769  & 0.7360  & 0.0807  & 0.7511  & \textbf{0.1214 } & \textbf{0.7714 } & 0.1155  \\
    \rowcolor[rgb]{ .851,  .851,  .851} Round Nut & 0.7110  & 0.0110  & 0.5280  & 0.0040  & 0.5470  & 0.0040  & 0.5340  & 0.0040  & 0.6314  & 0.0068  & 0.6196  & 0.0282  & 0.6658  & 0.0100  & 0.6948  & 0.0154  & 0.7105  & 0.0921  & 0.6582  & 0.0215  & \textbf{0.7136 } & \textbf{0.1061 } \\
    \midrule
    Mean  & 0.7017  & 0.0282  & 0.5340  & 0.0068  & 0.5565  & 0.0179  & 0.5417  & 0.0166  & 0.6426  & 0.0450  & 0.6686  & 0.0595  & 0.6697  & 0.0499  & 0.6989  & 0.0778  & 0.7164  & 0.0845  & 0.6957  & 0.0812  & \textbf{0.7252 } & \textbf{0.1089 } \\
    \bottomrule
    \bottomrule
    \end{tabular}}%
  \label{tab5}%
\end{table*}%

As indicated in $Eq.$ \ref{eq1}, the parameter $\theta_a$ of apprentice encoder is optimized after spatial discrepancy optimization. Finally, the anomaly score can be calculated by $\mathcal{A}: d \rightarrow \mathbb{R}$:

\begin{equation}
\label{eq5}
\mathcal{A}(d_k^-) = \parallel \Phi_e(d_k^-;\theta_e), \Phi_a(d_k^-;\theta_a) \parallel_2
\end{equation}

where $d_k^-$ represents the sample of testing data, $\theta_a$ is the optimized parameter after training. 

\section{Experiments}
\label{sec:experiment}

\subsection{Benchmark Construction}

\textbf{Experiment Setup.} All experiments are deployed on a computing platform equipped with Rocky Linux 8, which is equipped with an Xeon 4110 processor, 8 NVIDIA A800 GPUs (80GB), and 512GB of RAM. We followed the original parameters used in each algorithm. What’s more, The experimental results are taken as the average of five times.

\textbf{Evaluation Metrics.} An unsupervised AD task is measured using the Area Under the Receiver Operating Characteristic Curve (AUROC) and the Area Under the Precision-Recall Curve (AUPR) at the object and point levels. We follow the above metric to build our benchmark.

\textbf{Comparison Methods.} In order to ensure the rationality and fairness of benchmark construction, we ignore some methods related to RGBD data and instead focus on only targeting point cloud data. So we select several state-of-the-art methods to construct benchmark, such as BTF \cite{Horwitz_2023_CVPR}, M3DM \cite{wang2023multimodal}, PatchCore \cite{9879738}, Reg3D-AD \cite{DBLP:conf/nips/LiuXCLWLWZ23}, IMRNet \cite{10658462}, and R3D-AD \cite{zhou2024r3dad}. To further expand the richness of the validation algorithm, we introduce four modes, including Raw, FPFH, PointMAE \cite{DBLP:journals/pami/LiuT16}, and PointBERT \cite{DBLP:conf/cvpr/YuTR00L22}. Specifically, Raw represents adopting the coordinate features (XYZ) into the pipeline, FPFH introduces fast point feature histogram operation, and PointMAE and PointBERT is are two different point cloud feature extractors. We can obtain ten comparison experiments to validate the IEC3D-AD through different combinations. What's more, we also validated the performance of our method in IEC3D-AD to provide a new baseline for this dataset.

\begin{figure}
    \centering
    \includegraphics[width=\linewidth]{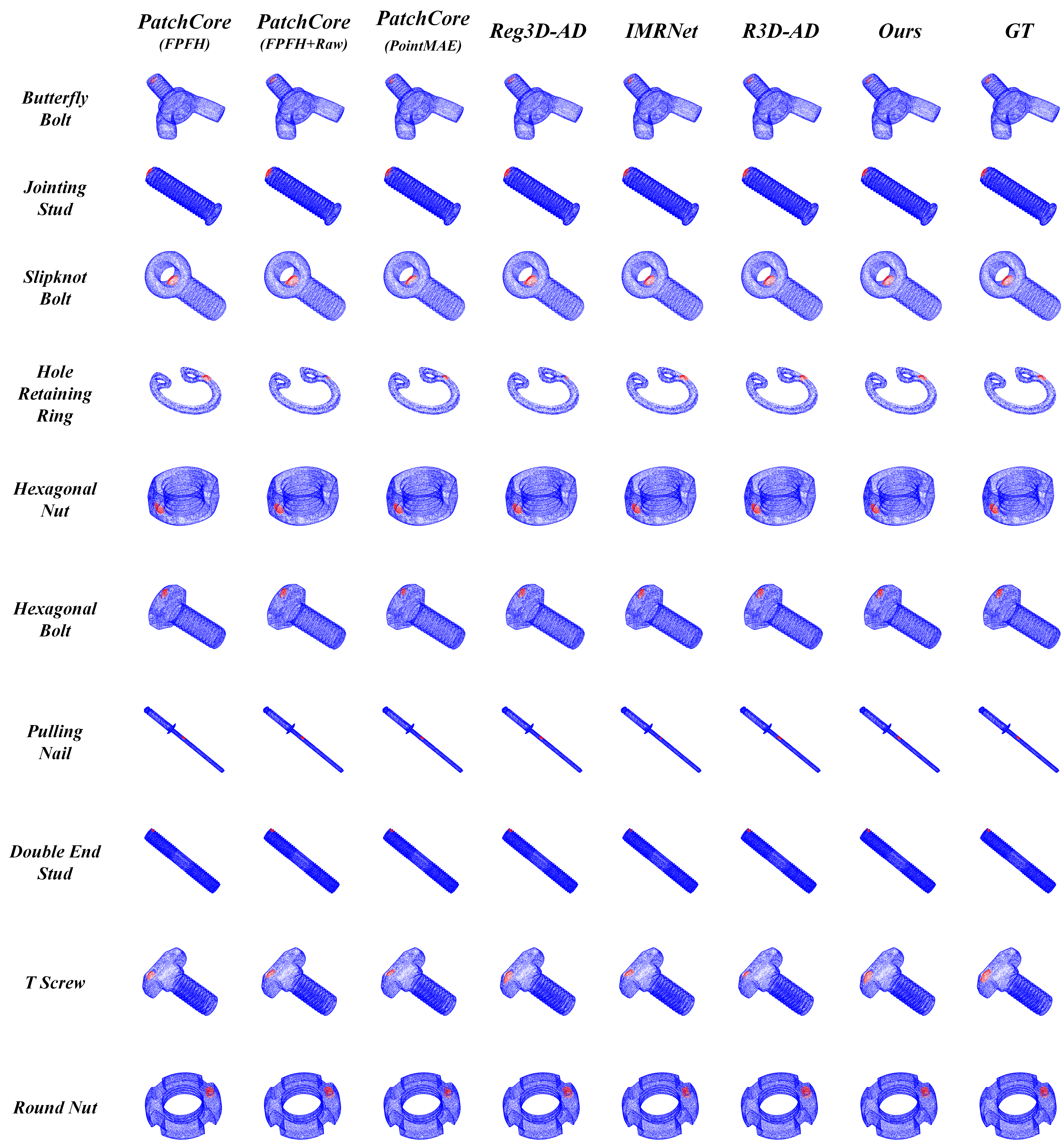}
    \caption{The visualization maps of different categories.}
    \label{fig5}
\end{figure}

\subsection{Comparison Experiments}
\textbf{Quantitative Results.} The evaluation under the object level and point level focuses on the abnormal situation of the entire object and details of local point clouds, respectively. The results under object level is shown in Tab. \ref{tab4}, and the results under point level is shown in Tab. \ref{tab5}. While our method could not get optimal results across all industrial equipment component types, the final mean result is superior to  previous state-of-the-art methods. For both AUROC and AUPR under object level, our method achieves the highest sc
ore for 10 out of 15 categories. For AUROC and AUPR under point level, our method achieved 11 highest scores and 9 highest scores respectively. However, at AUPR of the point level, the values of all algorithms cannot achieve the performance of other metric. This phenomenon is due to the imbalanced nature of the dataset, with outliers accounting for only 0.78\%-2.28\% of the overall point cloud, which is a low ratio. This indicates that in unsupervised AD task, it is easier to identify abnormal samples at the object level.

\begin{figure}
    \centering
    \includegraphics[width=0.9\linewidth]{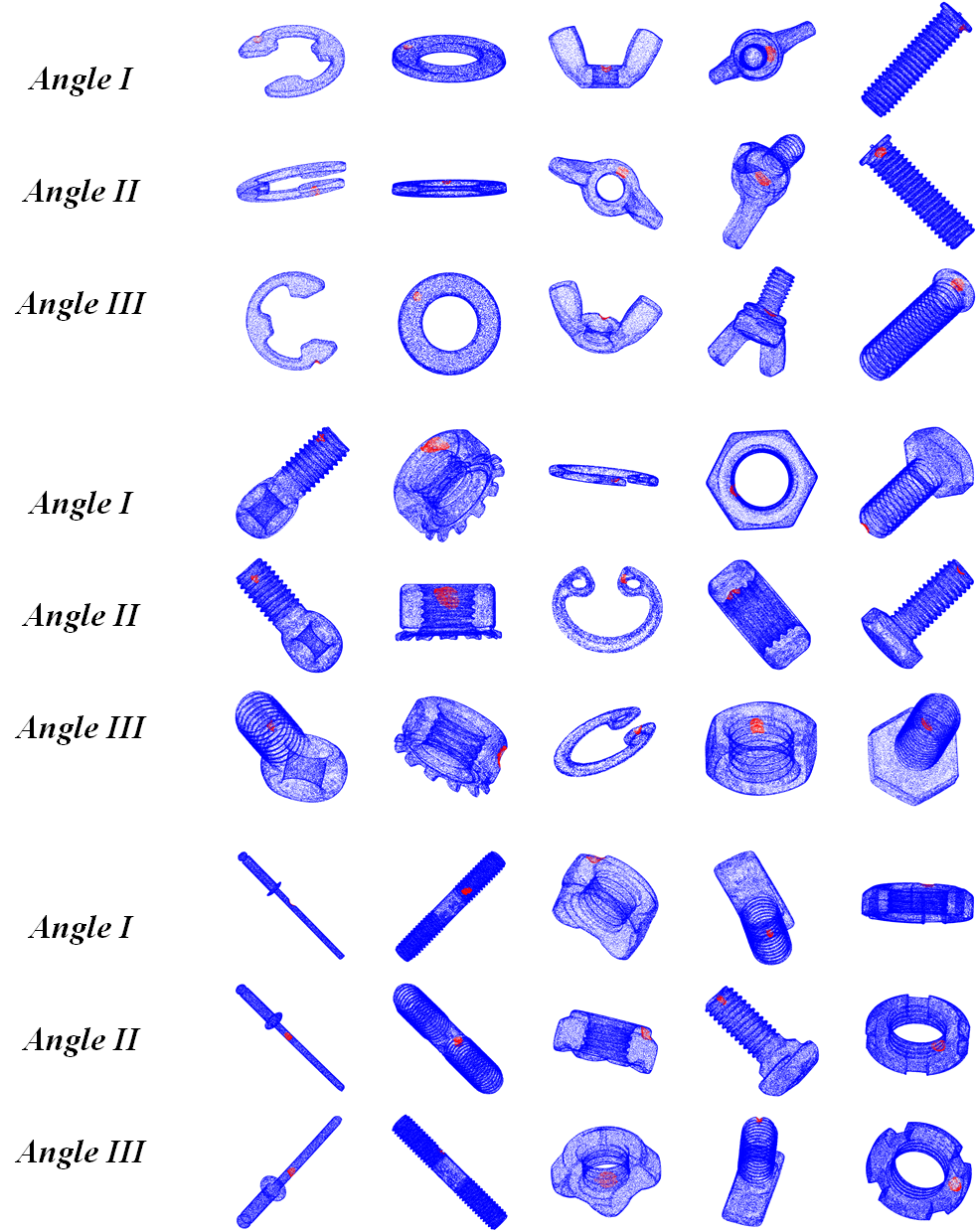}
    \caption{The visualization maps under different perspectives.}
    \label{fig8}
\end{figure}

\textbf{Qualitative Results.} We visualize the algorithm results in the benchmark to compare and analyze the qualitative results. It should be noted that due to the poor visualization results of BTF \cite{Horwitz_2023_CVPR} and M3DM \cite{wang2023multimodal}, we ignored them and only retained the visualization results of the remaining algorithms. The specific details are shown in Fig. \ref{fig5}. From the visualization results, it can be seen that our method performs better in identifying subtle edge areas, which can avoid overcategorizing normal points as abnormal points. Moreover, we presented prediction results from different perspectives to provide a better visualization, as shown in Fig. \ref{fig8}.

\subsection{Comparison of algorithm efficiency} 
Efficiency is critical for real-world deployment in 3D-AD task. As shown in Tab. \ref{tab:r1}, our performance is still acceptable. It is worth mentioning that we do not calculate computational cost because the pytorch versions of various algorithms are not consistent, which may lead to biased results.

\begin{table}[htbp]
  \centering
  \caption{Comparison of algorithm efficiency.}
  \resizebox{\linewidth}{!}{
    \begin{tabular}{c|cc}
    \toprule
    \toprule
    \multirow{2}[4]{*}{Method} & \multicolumn{2}{c}{Efficiency Metrics} \\
\cmidrule{2-3}          & Memory footprint (GB) & Inference Speed (FPS) \\
    \midrule
    BTF   & 2.82  & 1.21 \\
    M3DM  & 2.94  & 0.24 \\
    PatchCore & 4.41  & 0.12 \\
    Reg3D-AD & 6.18  & 0.06 \\
    IMRNet & 3.75  & 0.09 \\
    R3D-AD & \textbf{2.43} & \textbf{5.41} \\
    GMANet(Ours) & \underline{2.79}  & \underline{1.34} \\
    \bottomrule
    \bottomrule
    \end{tabular}}%
  \label{tab:r1}%
\end{table}%

\subsection{Ablation Studies}
\begin{figure}
    \centering
    \includegraphics[width=\linewidth]{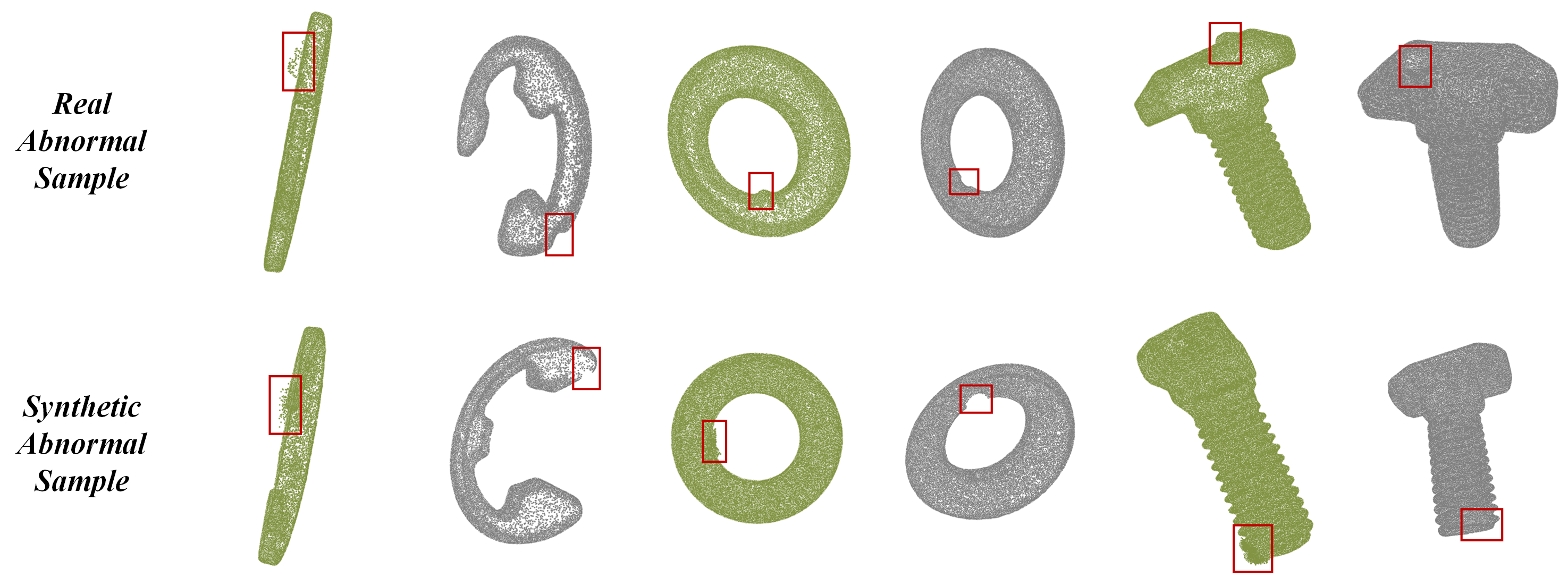}
    \caption{Comparisons between real and synthetic samples.}
    \label{fig3}
\end{figure}

\textbf{Module analysis.} To validate the impact of each module, we separately eliminate the SPCG and SDO modules. Above all, we replace SPCG with other adversarial attacking methods \cite{DBLP:conf/icip/LiuYS19,DBLP:journals/mms/LiangLNL22,DBLP:conf/iccv/ZhouCZFZY19,DBLP:conf/cvpr/Dong0Z0ZY20} to generate synthetic abnormal samples. Moreover, we assume that the normal features will not affect the abnormal features to remove SDO module. That means we only need to minimize the $d_n$ and neglect the $d_s$. As shown in Tab. \ref{tab6}, The results indicate that SPCG and SDO modules can improve AUROC by 8.81\% and AUPR by 40.76\%. Moreover, we also visualize the synthetic abnormal samples generated by SPCG, as shown in Fig. \ref{fig3}.

\begin{table}[htbp]
  \centering
  \caption{The results of different model variants.}
  \resizebox{\linewidth}{!}{
    \begin{tabular}{c|cc|cc|cc}
    \toprule
    \toprule
    \multirow{3}[4]{*}{Category} & \multicolumn{2}{c|}{\multirow{2}[2]{*}{w/o SPCG}} & \multicolumn{2}{c|}{\multirow{2}[2]{*}{w/o SDO}} & \multicolumn{2}{c}{\multirow{2}[2]{*}{GMANet (Ours)}} \\
          & \multicolumn{2}{c|}{} & \multicolumn{2}{c|}{} & \multicolumn{2}{c}{} \\
\cmidrule{2-7}          & AUROC & AUPR & AUROC & AUPR & AUROC & AUPR \\
    \midrule
    \rowcolor[rgb]{ .851,  .851,  .851} Fender Ring & 0.7531  & 0.0916  & 0.7621  & 0.0821  & 0.8386  & 0.1397  \\
    Washer & 0.7735  & 0.1202  & 0.7692  & 0.1598  & 0.8808  & 0.2981  \\
    \rowcolor[rgb]{ .851,  .851,  .851} Butterfly Nut & 0.7232  & 0.0452  & 0.7289  & 0.0592  & 0.7712  & 0.0643  \\
    Butterfly Bolt & 0.7560  & 0.2371  & 0.7644  & 0.2611  & 0.7797  & 0.3140  \\
    \rowcolor[rgb]{ .851,  .851,  .851} Joingting Stud & 0.4072  & 0.0131  & 0.4862  & 0.0256  & 0.5932  & 0.0360  \\
    Slipknot Bolt & 0.5289  & 0.0211  & 0.5672  & 0.0287  & 0.5906  & 0.0331  \\
    \rowcolor[rgb]{ .851,  .851,  .851} K Nut & 0.7453  & 0.0732  & 0.7882  & 0.0792  & 0.8273  & 0.0968  \\
    Hole Retaining Ring & 0.7081  & 0.0289  & 0.7806  & 0.0423  & 0.8160  & 0.0557  \\
    \rowcolor[rgb]{ .851,  .851,  .851} Hexagonal Nut & 0.6982  & 0.0402  & 0.6641  & 0.0511  & 0.6551  & 0.0604  \\
    Hexagonal Bolt & 0.5621  & 0.0098  & 0.5589  & 0.0092  & 0.6182  & 0.0114  \\
    \rowcolor[rgb]{ .851,  .851,  .851} Pulling Nail & 0.6128  & 0.1071  & 0.6023  & 0.1021  & 0.6504  & 0.1272  \\
    Double End Sutd & 0.5932  & 0.0903  & 0.6079  & 0.1018  & 0.6470  & 0.1121  \\
    \rowcolor[rgb]{ .851,  .851,  .851} Square Welded Nut & 0.6674  & 0.0513  & 0.6791  & 0.0548  & 0.7245  & 0.0627  \\
    T Screw & 0.6986  & 0.0821  & 0.7032  & 0.0924  & 0.7714  & 0.1155  \\
    \rowcolor[rgb]{ .851,  .851,  .851} Round Nut & 0.6387  & 0.0735  & 0.6682  & 0.0985  & 0.7136  & 0.1061  \\
    \midrule
    \multirow{2}[2]{*}{Mean} & \multirow{2}[2]{*}{0.6578 } & \multirow{2}[2]{*}{0.0723 } & \multirow{2}[2]{*}{0.6754 } & \multirow{2}[2]{*}{0.0832 } & \multicolumn{1}{c}{\multirow{2}[2]{*}{0.7252 ($8.81\% \uparrow$)}} & \multicolumn{1}{c}{\multirow{2}[2]{*}{0.1089 ($40.76\% \uparrow$)}} \\
          &       &       &       &       &       &  \\
    \bottomrule
    \bottomrule
    \end{tabular}}%
  \label{tab6}%
\end{table}%

\begin{table}[htbp]
  \centering
  \caption{Evaluation of different empirical parameters.}
  \resizebox{\linewidth}{!}{
    \begin{tabular}{c|c|cc}
    \hline
    \hline
    \multicolumn{2}{c|}{\multirow{2}[4]{*}{Metric}} & \multicolumn{2}{c}{Empirical Parameters} \\
\cline{3-4}    \multicolumn{2}{c|}{} & Curvature threshold 0.05/0.10/0.15/0.20 & Point ratio 0.05/0.10/0.15/0.20 \\
    \hline
    \multirow{2}[2]{*}{Object level} & AUROC & 0.8012/0.8321/0.8487/0.8798 & 0.7892/0.8093/0.8767/0.8215 \\
          & AUPR  & 0.8113/0.8322/0.8481/0.8601 & 0.8113/0.8322/0.8601/0.8481 \\
    \hline
    \multirow{2}[2]{*}{Point level} & AUROC & 0.6172/0.6461/0.6631/0.7109 & 0.6342/0.6615/0.7097/0.6712 \\
          & AUPR & 0.0507/0.0511/0.0882/0.1081 & 0.0555/0.0611/0.1075/0.0911 \\
    \hline
    \hline
    \end{tabular}}%
  \label{r2}%
\end{table}%

\textbf{Empirical parameter analysis.}  (1) The values of the empirical parameters were obtained through multiple experiments. As shown in Tab. \ref{r2}, the values we have chosen have been consistently optimal in multiple comparisons and their combination is also optimal. (2) $\omega$ in Eq. 3 and 4 are calculated from the extracted feature distribution and are not a fixed value. $\alpha$ is set to 0.01.

\subsection{Performance on Other Datasets}
In order to further verify the effectiveness of our algorithm, we also evaluate its performance on Real3D-AD and Anomaly-ShapeNet, as illustrated in Tab. \ref{tab7}. The percentage represents the higher or lower proportion than the SOTA algorithm in the original paper. Due to the missing source code of Anomaly-ShapeNet, we can only compare the public results. Specifically, our algorithm has excellent performance on Real3D-AD and Anomaly-ShapeNet data with the same defect type. The algorithm also performs well on samples with non-similar defect types. 

\begin{table}[htbp]
  \centering
  \caption{Performance on other datasets.}
  \resizebox{\linewidth}{!}{
    \begin{tabular}{c|c|c|ccc}
        \toprule
        \toprule
        \multicolumn{2}{c|}{Metric} & \multicolumn{4}{c}{Dataset} \\
        \cline{3-6} \multicolumn{2}{c|}{} & RealAD-3D & \multicolumn{3}{c}{Anomaly-ShapeNet} \\
        \cline{3-6} \multicolumn{2}{c|}{} & Similar Defect Category & Similar Defect Category & Non-similar Defect Category & AVG. \\
        \midrule
        \multirow{2}[2]{*}{Object level} & AUROC & 0.741 (5.25\% $\uparrow $) & 0.677 (2.58\% $\uparrow $) & 0.669 (1.05\% $\uparrow $) & 0.673 (1.32\% $\uparrow $) \\
        & AUPR & 0.756 (4.56\% $\uparrow $) & 0.628 (2.11\% $\uparrow $) & 0.632 (1.76\% $\downarrow $) & 0.630 (1.41\% $\uparrow $) \\
        \midrule
        \multirow{2}[2]{*}{Point level} & AUROC & 0.755 (7.85\% $\uparrow $) & 0.685 (4.90\% $\uparrow $) & 0.653 (0.93\% $\uparrow $) & 0.689 (3.14\% $\uparrow $) \\
        & AUPR & 0.147 (30.09\% $\uparrow $) & - & - & - \\
        \bottomrule
        \bottomrule
    \end{tabular}}
  \label{tab7}
\end{table}%

\begin{table}[htbp]
  \centering
  \caption{The result of fully supervised task.}
    \resizebox{\linewidth}{!}{
    \begin{tabular}{c|c|ccc|cc|cc}
    \toprule
    \toprule
    \multirow{2}[4]{*}{Methods} & \multirow{2}[4]{*}{Publication \& Year} & \multicolumn{3}{c|}{Average Metrics} & \multicolumn{2}{c|}{Normal Metrics} & \multicolumn{2}{c}{Abnormal Metrics} \\
\cmidrule{3-9}          &       & mIoU  & mAcc  & allAcc & iou   & accuracy & iou   & accuracy \\
    \midrule
    pointnet++ & NeurIPS 2017 & 0.7846 & 0.8166 & 0.9942 & 0.9942 & 0.9987 & 0.5750 & 0.6344 \\
    pointnet++msg & NeurIPS 2017 & 0.8068 & 0.8352 & 0.9949 & 0.9948 & 0.9989 & 0.6187 & 0.6714 \\
    spunet & CVPR 2019 & 0.8823 & 0.9081 & 0.9969 & 0.9969 & 0.9992 & 0.7678 & 0.8170 \\
    PTv1  & ICCV 2021 & 0.8756 & 0.9053 & 0.9967 & 0.9967 & 0.9991 & 0.7544 & 0.8115 \\
    pointnext-l & NeurIPS 2022 & 0.8431 & 0.8941 & 0.9967 & 0.9956 & 0.9982 & 0.6906 & 0.7900 \\
    octformer & ACM TG 2023 & 0.8498 & 0.8896 & 0.9959 & 0.9959 & 0.9986 & 0.7037 & 0.7805 \\
    pointvecoter-l & CVPR 2023 & 0.8586 & 0.9079 & 0.9961 & 0.9961 & 0.9983 & 0.7212 & 0.8174 \\
    oacnns & CVPR 2024 & 0.8710 & 0.8961 & 0.9967 & 0.9966 & 0.9992 & 0.7454 & 0.7930 \\
    \bottomrule
    \bottomrule
    \end{tabular}}%
  \label{tab:s2}%
\end{table}%

\subsection{Fully supervised AD benchmark} 

Except unsupervised AD benchmark, we also conduct fully supervised AD benchmark. Due to the use of labeled data during training process, fully supervised methods are typically able to more accurately identify and classify anomalies. However, obtaining large-scale annotated datasets may be costly, and models may overfit the annotated data and lack sufficient detection capabilities for unseen types of anomalies. 

In fully supervised anomaly detection task, we only utilize the abnormal samples $\mathbb{D}$ in IEC3D-AD dataset. During the training phase, we partition the data into training and testing sets using an 8:2 ratio. Then we calculate the average of numerous test results to ensure the stability of the results. Fully supervised anomaly detection task is measured using accuracy (ACC) and mean Intersection over Union (mIoU). ACC represents the proportion of correctly classified points in the algorithm to the total number of points. mIoU is the overlap area between the predicted and ground truth points. We will evaluate the metrics of normal and abnormal samples separately and calculate the average metrics.

We follow the pipeline of Pointcept and verify each method on IEC3D-AD. However, there are only eight methods achieving excellent performance. The details are shown in Tab. \ref{tab:s2}. We only listed the algorithm results that performed well, and we did not list the other results that performed poorly. Unlike previous 3D datasets, our data comes from samples in real industrial scenes, with a much smaller proportion of anomalies compared to carefully designed samples in laboratory environments. Therefore, some existing 3D fully supervised algorithms cannot achieve good results in this scenario, which indirectly indicates that there is a significant difference between real scene data and idealized data. Among the algorithms we have listed, PTV1 achieves the best performance on average metrics, and spunet exhibits superior performance on abnormal metrics.

\section{Conclusion}
\label{sec:conclusion}
In this paper, we propose a novel IEC3D-AD dataset for the industrial equipment components anomaly detection task. IEC3D-AD is compiled from authentic industrial equipment components and has a smaller proportion of defects, thereby imparting significant research merit. We propose a comprehensive benchmark for 3D anomaly detection under unsupervised AD scenarios. This benchmark evaluates the performance of the dataset. In addition, we introduce a novel 3D-AD paradigm GMANet which generates synthetic point cloud samples based on geometric morphological analysis. And GMANet reduces the margin and increases the overlap between normal and abnormal point-level features through spatial discrepancy optimization. Extensive experiments demonstrate the effectiveness of our method.

\bibliographystyle{IEEEtran}
\bibliography{ref}

\begin{IEEEbiographynophoto}{Bingyang Guo} is currently a PhD candidate in the software college of Northeastern University China. He received his Bachelor degree in mechanical engineering from Shenyang Ligong University, and Master degree in mechanical design and theory from Northeastern University, China. His research focuses on image segmentation and defect detection.
\end{IEEEbiographynophoto}

\begin{IEEEbiographynophoto}{Hongjie Li} is currently a master's student in the software college of Northeastern University China. He holds a Bachelor's degree in network engineering from Inner Mongolia University. His research focuses on point cloud segmentation and point cloud defect detection.
\end{IEEEbiographynophoto}

\begin{IEEEbiographynophoto}{Ruiyun Yu} is currently a professor and vice dean of the Software College at the Northeastern University, China. He received his Ph.D. and M.S. degree in computer science and bachelor degree in mechanical engineering from the Northeastern University in 2009, 2004, and 1997, respectively. He serves as the director of center for Cross-media Artificial Intelligence. He is one of the Baiqianwan Talents of Liaoning Province, China (Hundred Talents Level), and now a member of the CCF IoT Committee, and a Senior Member of CCF. His research interests include intelligent sensing and computing, computer vision, data intelligence. 
\end{IEEEbiographynophoto}

\begin{IEEEbiographynophoto}{Hanzhe Liang} is currently pursuing dual degrees, a Bachelor of Management in Information Management and Information Systems at Shenzhen University, Shenzhen, China, and a Bachelor of Science at Audenica Business School. His research interests include computer vision, AI4Science, and machine learning.
\end{IEEEbiographynophoto}

\begin{IEEEbiographynophoto}{Jinbao Wang} received his Ph.D. degree from the University of Chinese Academy of Sciences in 2019. He is currently an Assistant Professor with the School of Artificial Intelligence, Shenzhen University, Shenzhen, China. His research interests include digital human modeling and driving, image anomaly detection, computer vision, and machine learning.
\end{IEEEbiographynophoto}

\end{document}